\documentclass[sigconf,authorversion]{acmart}

\AtBeginDocument{%
  }

\copyrightyear{2024}
\acmYear{2024}
\setcopyright{acmlicensed}\acmConference[ISWC '24]{Proceedings of the 2024 ACM International Symposium on Wearable Computers}{October 5--9, 2024}{Melbourne, VIC, Australia}
\acmBooktitle{Proceedings of the 2024 ACM International Symposium on Wearable Computers (ISWC '24), October 5--9, 2024, Melbourne, VIC, Australia}
\acmDOI{10.1145/3675095.3676623}
\acmISBN{979-8-4007-1059-9/24/10}

\usepackage{xspace}
\usepackage{algorithm} 
\usepackage{algpseudocode}
\usepackage{makecell}

\begin{document}

\title{RetailOpt: Opt-In, Easy-to-Deploy Trajectory Estimation from Smartphone Motion Data and Retail Facility Information}

\author{Ryo Yonetani}
\email{yonetani_ryo@cyberagent.co.jp}
\orcid{0000-0002-2724-6233}
\affiliation{%
  \institution{CyberAgent}
  \streetaddress{40-1 Utagawa-cho, Shibuya-ku}
  \city{Tokyo}
  \country{Japan}
  \postcode{150-0042}
}

\author{Jun Baba}
\email{baba_jun@cyberagent.co.jp}
\orcid{0000-0003-0680-5021}
\affiliation{%
  \institution{CyberAgent}
  \streetaddress{40-1 Utagawa-cho, Shibuya-ku}
  \city{Tokyo}
  \country{Japan}
  \postcode{150-0042}
}

\author{Yasutaka Furukawa}
\email{furukawa@sfu.ca}
\orcid{0009-0006-9775-4512}
\affiliation{%
  \institution{Simon Fraser University}
  \streetaddress{8888 University Drive}
  \city{Burnaby, BC}
  \country{Canada}
  \postcode{V5A 1S6}
}

\def\eg{{\it e.g.}}
\def\cf{{\it c.f.}}
\def\ie{{\it i.e.}}
\def\etal{{\it et al. }}
\def\etc{{\it etc}}

\newcommand{\method}{\textsc{RetailOpt}\xspace}
\newcommand{\Xtu}{\mathcal{X}_\mathrm{tu}}
\newcommand{\Xtk}{\mathcal{X}_\mathrm{tk}}
\newcommand{\Oobs}{\Omega_\mathrm{obs}}

\renewcommand{\shortauthors}{Ryo Yonetani, Jun baba, and Yasutaka Furukawa}

\begin{abstract}
We present RetailOpt, a novel opt-in, easy-to-deploy system for tracking customer movements offline in indoor retail environments. The system uses readily accessible information from customer smartphones and retail apps, including motion data, store maps, and purchase records. This eliminates the need for additional hardware installations/maintenance and ensures customers full data control. Specifically, RetailOpt first uses inertial navigation to recover relative trajectories from smartphone motion data. The store map and purchase records are cross-referenced to identify a list of visited shelves, providing anchors to localize the relative trajectories in a store through continuous and discrete optimization. We demonstrate the effectiveness of our system in five diverse environments. The system, if successful, would produce accurate customer movement data, essential for a broad range of retail applications including customer behavior analysis and in-store navigation.
\end{abstract}




\maketitle

\section{Introduction}
\label{sec:introduction}

Imagine owning a fashion store. You would know the locations of entrances, walls, and shelves, and what items are displayed on which shelves. The purchase records from the points-of-sales (POS) cash registers reveal which customer purchased which items. Such granular retail facility information provides rich constraints for tracking the movement of every customer, holding great potential for retail applications such as targeted marketing~\cite{liu2018tar,ghose2019mobile,schwipper2020mobile}, in-store navigation~\cite{kamei2011effectiveness,purohit2013sugartrail,paolanti2018modelling}, inventory management~\cite{carreras2013store}, and anomalous behavior detection~\cite{morais2019learning,rodrigues2020multi,belhadi2020trajectory}. Indoor people tracking technologies could impact broader domains beyond the retail industry, for example, enhancing visitor experiences in museum or theme parks~\cite{kontarinis2017towards} and assistive navigation technologies~\cite{ohnbar2018variability}. 

Location-based services should require minimal additional costs and employ an opt-in mechanism to respect privacy. Existing indoor localization or motion-tracking systems require an expensive site survey to collect referential environmental information, such as visual features~\cite{liang2013image,walch2017image,sattler2011fast,piasco2018survey,toft2020long,naseer2018robust}, wireless fingerprints~\cite{kaemarungsi2004modeling,honkavirta2009comparative,faragher2015location,vo2015survey,nessa2020survey,zhao2023nerf,sugasaki2017robust,zafari2019survey}, or magnetometer responses~\cite{xie2014maloc,li2015using}, at every location within the environment. The collected information requires maintenance upon environmental changes. Active installation of hardware components, such as ultra-wideband (UWB) systems~\cite{ridolfi2021self} or surveillance cameras~\cite{brunetti2018computer,hu2004survey,valera2005intelligent,gawande2020pedestrian}, achieves high precision and is more resilient to environmental changes. Still, hardware installation beyond existing infrastructure incurs significant costs for building/store owners. Despite their prevalence, security camera systems in retail settings do not inherently support privacy safeguards, as they capture images of anyone regardless of consent.

\begin{figure}
    \centering
\includegraphics[width=\linewidth]{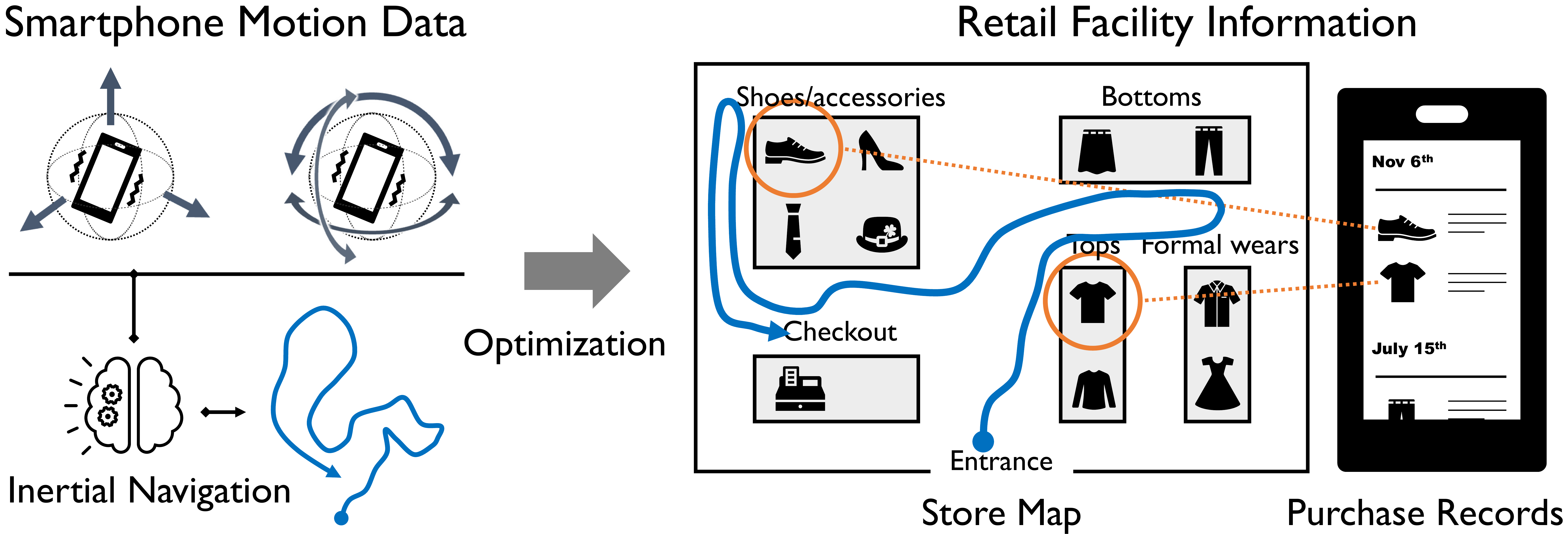}
    \caption{The \method system leverages smartphone motion data, store map, and purchase records to enable opt-in, easy-to-deploy indoor trajectory estimation.}
    \label{fig:overview}
\end{figure}

\begin{figure*}[t]
    \centering
\includegraphics[width=\linewidth]{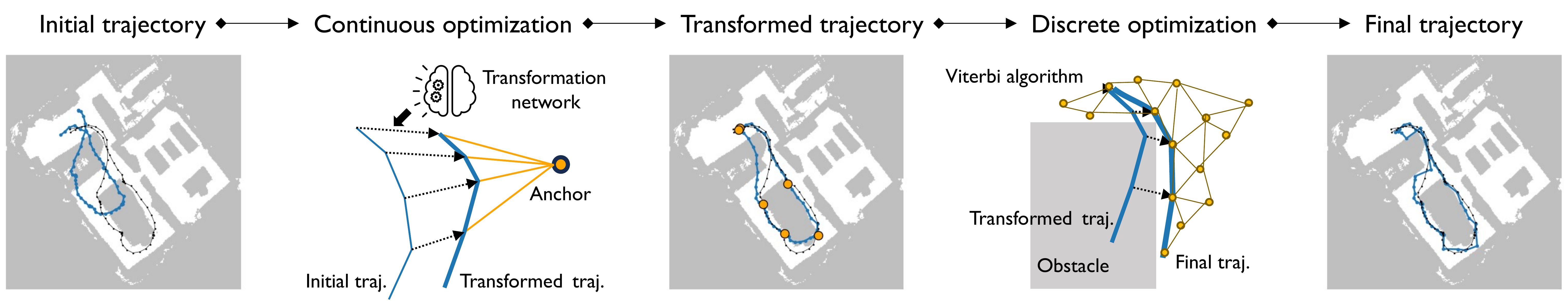}
    \caption{Trajectory estimation pipeline. A relative motion trajectory from inertial navigation (blue lines in the figure, while ground-truth trajectories obtained by SLAM are shown in black lines) is first transformed to better align with anchors (orange circles) created from the store map and purchase records. The transformed trajectory is then projected onto the valid space using a Viterbi algorithm, ensuring the final trajectory does not collide with obstacles in the environment.} 
    \label{fig:optim}
\end{figure*}

This paper proposes \method, a novel system that utilizes smartphones as a wearable sensor to track customer movements in indoor retail environments. Our system fuses two information: the smartphone motion data and retail facility information (see Fig.~\ref{fig:overview}). Inertial navigation recovers a relative motion trajectory of each customer from inertial measurement unit (IMU) sensor data~\cite{yan2020ronin}. Retail facility information, specifically \emph{store maps and purchase records}, identify \emph{anchors} that indicate global reference positions and help localize the relative trajectory in a store. The store map provides the physical layout (\ie, locations of shelves, corridors, doors, \etc.) and the semantic layout (\ie, ``which items are on which shelves.'') The purchase records (\ie, ``which items were purchased'') reveal which shelves were visited, thereby serving as anchors that reference their locations. The main technical challenge is how to align relative trajectories with the anchors \emph{without knowing when they were visited}. As shown in Fig.~\ref{fig:optim}, our novel continuous-discrete optimization pipeline matches relative trajectories to time-unknown anchors via gradient descent and projects them onto the non-obstacle space via the Viterbi algorithm.

The \method system requires minimal additional costs for hardware installation and maintenance. Modern smartphones are equipped with IMUs that are easily accessible from third-party apps with permission from users. Arguably, a large number of people use retailer's smartphone apps to access store information, check sale items, or earn reward points~\cite{molinillo2022customer,van2019engaging,wohllebe2020mobile,kang2015store}. Recent apps provide customer's purchase histories as well as in-store maps.\footnote{Walmart app is a good example: \url{https://www.walmart.com/cp/find-an-item-store-maps/6422847}.} Such facility information is kept up-to-date in retailer's daily operations, thus minimizing installation/maintenance costs compared to existing localization systems introduced earlier. Furthermore, our system enables customer tracking through opt-in consent. Unlike visual tracking and other passive localization techniques, we estimate trajectories using data from customers' smartphones, ensuring customers have full control over their data.

We evaluate the \method system in five diverse environments. Our system archives considerably lower average positional error than baseline methods~\cite{yan2020ronin,golyanik2016extended,hirose2020bayesian} and  outperforms a state-of-the-art inertial localization method~\cite{herath2022niloc}.

\section{Related Work}
\label{sec:related_work}

\paragraph{Tracking People using IMUs} Inertial navigation and pedestrian dead reckoning (PDR) utilize acceleration and angular velocity measurements from IMUs to estimate the relative positions of subjects, that is, positions relative to the initial position~\cite{wu2019survey,hou2020pedestrian}. To this end, supervised learning approaches collect pairs of IMU data and motion trajectories and learn a mapping function between them, often by a neural network model. The robust neural inertial navigation (RoNIN) system~\cite{yan2020ronin} is a landmark study that employs deep neural networks to enhance performances in challenging real-world scenarios. Neural inertial odometry combines neural network predictions with existing odometry systems, while assuming that IMUs are rigidly attached to the body~\cite{sun2021idol,liu2020tlio,saha2022tiny}. Building on these advancements, our focus is on how to geo-localize such relative trajectories from inertial navigation within the environment, aided by information from retail facilities. Neural inertial localization~\cite{herath2022niloc} is one of the most recent machine learning-based attempts to localize positions solely from IMU data.

\paragraph{Localization using Anchors}
Our system utilizes retail facility information to form anchors, \ie, referential landmarks with known global positions, to geo-localize relative trajectories from inertial navigation. Such anchors can be constructed using dedicated devices~\cite{li2015bluetooth,zhou2023gps,venkatnarayan2019enhancing,mahmoud2022ultra,carreras2013store,diallo2019wireless,kim2021deep,gummeson2017rfid,herath2021fusion}, other people’s smartphones~\cite{ding2022smart} or environmental contexts such as acoustic signals~\cite{tarzia2011indoor} and motion signals~\cite{wang2012no,purohit2013sugartrail}. Our anchor can be categorized as environmental context-based, which not only eliminates the need for additional device installation, but also ensures they remains updated during routine operations without requiring extra site surveys.

\section{Preliminaries}
\label{sec:preliminaries}
This section provides the mathematical definition of our trajectory estimation task. We consider a bounded 2D environment $\Omega=[0, 1]^2$. The trajectory of a target person is defined over a discrete time interval $[1, T]$ and represented by a sequence $P=(p_1,\dots,p_T)$, where $p_t\in\Omega$ is the position of the person at time $t$. 

\paragraph{Spatial Cues from Retail Facility Information}
We consider three types of spatial cues, including anchors created from store maps and purchase records for a target person.
\begin{itemize}
\item
Time-unknown anchors: $\Xtu=\{(x_j, \phi)\}_{j=1}^J,\;x_j\in\Omega$, provide a small set of locations that the target person has certainly visited, although \emph{it remains unknown when these visits happened}.\footnote{Here, $\phi$ indicates that no time information is available for $x_j$, making the notation of $\Xtu$ consistent with that of $\Xtu$.} For example, if an apple is listed in the target person's receipt, it is highly probable that the person visited the shelf of apples in the fruit section, whose location is available in the store map.
\item Time-known anchors: $\Xtk=\{(x_k, t_k)\}_{k=1}^K,(x_k, t_k)\in\Omega\times \mathbb{N}$, present a pair of locations and timestamps when these locations are visited. Though less frequently than time-unknown anchors, time-known anchors are also available for practical retail scenarios, \eg, when making payments at a POS register system.
\item Obstacle layout: $\Oobs \subset \Omega$, indicates locations and sizes of physical obstacles such as walls or pillars in the environment, which can be easily extracted from the store map.
\end{itemize}

\paragraph{Relative Trajectory from Smartphone's Motion Data}
Furthermore, the target person's smartphone provides continuous access to device acceleration, angular rate, and orientation data from its IMU. These data are utilized to estimate a relative motion trajectory of the target via inertial navigation, which we denote as $Q=(q_1,\dots,q_T), \; q_t\in \mathbb{R}^2$. In line with prior work~\cite{yan2020ronin,herath2022niloc}, we assume that the smartphone device is not necessarily attached rigidly to the person's body but is held loosely in hand, in a pocket, or in a bag. Any inertial navigation algorithm would essentially work as long as it offers reasonable motion estimation accuracy.

\paragraph{Key Technical Challenge}
Relative trajectories from smartphone's IMU are obtained densely but can be erroneous and drifted due to inherent inertial noises. On the other hand, anchors indicate actual global locations but are mostly unknown in time and available only sparsely for each journey (\eg, less than ten for hundreds of meters of movement). This leads to our key challenge: \emph{How can we integrate $Q$ and $\Xtu, \Xtk, \Oobs$ to estimate $P$?}

\section{\method System}
\label{sec:proposed_method}

\method system estimates a trajectory $P$ for a target person from the relative motion trajectory $Q$, time-unknown anchors $\Xtu$, time-known anchors $\Xtk$, and the obstacle layout $\Oobs$.
As outlined in Fig.~\ref{fig:optim}, \method involves continuous and discrete optimization procedures. Initially, we transform $Q$ to align more closely with actual locations indicated by $\Xtu$ and $\Xtk$. We achieve this by optimizing a neural network, hereafter referred to as the \emph{transformation network}, through gradient descent with back-propagation. Subsequently, we further refine the transformed trajectory by projecting it onto the valid space, \ie, $\Omega\setminus \Oobs$, based on a Viterbi algorithm. This ensures that the final trajectory does not collide with obstacles.

\subsection{Continuous Optimization}
Our transformation network accepts $Q$, $\Xtu, \Xtk$, and time indices $[1,\dots, T]$ to provide per-step displacements, $\Delta=(\delta_1,\dots,\delta_T)$, $\delta_t\in\mathbb{R}^2$, such that a transformed trajectory $Q'=(q'_1,\dots,q'_T),\; q'_t=q_t + \delta_t$ passes through the locations in $\Xtu, \Xtk$. The use of highly expressive neural networks enables us to derive a complex non-linear transformation while considering the spatiotemporal continuity of $Q$ as well as optimization targets indicated by $\Xtu, \Xtk$. Moreover, its optimization can be readily implemented on sophisticated deep learning frameworks with flexible automatic differentiation functionalities, allowing for efficient execution on GPUs.

Let us first define the objective function for optimizing the transformation network. The function measures the discrepancy between $Q'$ and $\Xtu, \Xtk$ as follows:
\begin{equation}
    L(Q', Q, \Xtu, \Xtk) = L_\mathrm{match}(Q', \Xtu, \Xtk) + L_\mathrm{bound}(Q') + \lambda L_\mathrm{reg}(Q', Q),
    \label{eq:objective}
\end{equation}
where $L_\mathrm{match}$ is a matching term, $L_\mathrm{bound}$ is a boundary constraint, and $L_\mathrm{reg}$ is a regularizer with coefficient $\lambda$, each detailed below.

\paragraph{Matching Term}
For each combination of a single anchor $(x,\cdot) \in \Xtu \cup \Xtk$ and transformed position $q'_t$ at time $t$, we evaluate the squared distance, $(q'_t - x)^2$, as well as a matching confidence $w_{(q'_t, t, x)}$ quantifying how likely $q'_t$ is to be actually located at $x$ at time $t$. In the matching term, $L_\mathrm{match}$, we take into account the matching that yields the highest value for the product of the discrepancy and confidence for each time $t$, defined as follows:
\begin{equation}
    L_\mathrm{match}(Q', \Xtu, \Xtk) = \sum_t \left(\max_{(x, \cdot) \in \Xtu\cup \Xtk} w_{(q'_t, t, x)} (q'_t - x)^2\right).
\end{equation}
If $x$ is from time-known anchors, the matching confidence $w_{(q'_t, t, x)}$ is given as $w_{(q'_t, t, x)} = 1$ if $(x, t') \in \Xtk,\; t' = t$ and $0$ otherwise, producing the highest confidence if and only if $t$ is in $\Xtk$. Evaluating $w_{(q'_t, t, x)}$ for time-unknown anchors is non-trivial because it is unknown at which time $t$ we should match $q'_t$ against $x$. To address such temporal uncertainty, we propose a `soft'-matching:
\begin{equation}
    w_{(q'_t, t, x)} = \frac{\exp (-(q'_t - x)^2 / \tau)}{\sum_t \exp(-(q'_t - x)^2 / \tau)},\; \mathrm{if} (x, \cdot)\in \Xtu.
    \label{eq:weight}
\end{equation}
This is a soft-max operation on the negative of squared distance $(q'_t - x)^2$ with temperature $\tau$, providing a higher confidence weight for pairs $(q'_t, x)$ that are closer to each other.

\paragraph{Boundary Constraint and Regularizer} The boundary constraint, $L_\mathrm{bound}(Q')$, ensures that the transformed position $q'_t$ remains within the bounded 2D space $\Omega=[0, 1]^2$, defined as follows:
\begin{equation}
L_\mathrm{bound}(Q') = \max_t \left(\mathrm{ReLU}(-q'_t) + \mathrm{ReLU}(q'_t-1)\right)^2,
\end{equation}
where $\mathrm{ReLU}(\cdot)$ is a rectified linear unit (ReLU) function that imposes a penalty for positions outside $[0, 1]^2$. Finally, we introduce a L1 regularizer for motion velocities:
\begin{equation}
    L_\mathrm{reg}(Q', Q)=\sum_t\left\| (q'_t - q'_{t-1}) - (q_t - q_{t-1})\right\|=\sum_t \left\|\delta_t - \delta_{t-1}\right\|,
\end{equation}
which prevents the motion velocity $q'_t - q'_{t-1}$ from changing too much from the original one $q_t - q_{t-1}$ in every step.

\paragraph{Network Architecture}
We design the architecture of the transformation network as follows:
\begin{equation}
\begin{aligned}
    H&=(h_1,\dots,h_T)=\mathrm{attn}(Q^\top W_\mathrm{q}, XW_\mathrm{k}, XW_\mathrm{v}),\\
    z_t&=\mathrm{cat}(q_t, h_t, t),\; \delta_t = \mathrm{MLP}(z_t).
    \label{eq:network}
\end{aligned}
\end{equation}
$X \in \Omega^{J+K}$ is a stack of all locations contained in $\Xtu$ and $\Xtk$. $\mathrm{attn}(\cdot,\cdot,\cdot)$ is a scaled dot-product attention layer~\cite{vaswani2017attention} with $W_\mathrm{q}, W_\mathrm{k}, W_\mathrm{V}\in\mathbb{R}^{2\times d}$ and $H\in\mathbb{R}^{T\times 2}$. $\mathrm{cat}(\cdot,\cdot,\cdot)$ is a concatenation of vectors. $\mathrm{MLP}$ is a standard multi-layer perceptron with the ReLU activation for each layer except the last one. The use of the attention layer allows the network architecture to flexibly accommodate trajectories $Q$ and anchors $X$ of arbitrary lengths within the same framework.

\paragraph{Optimization Strategy}
Importantly, all the terms in Eq.~(\ref{eq:objective}) as well as network layers in Eq.~(\ref{eq:network}) are differentiable with respect to per-step displacements, $\delta_t = q'_t - q_t$. This makes it possible to adopt a standard gradient descent technique to minimize the objective with respect to $\delta_t$, and consequently with respect to the parameters of the transformation network via back-propagation. Furthermore, the temperature parameter, $\tau$ in Eq.~(\ref{eq:weight}), can also be optimized simultaneously through gradient descent.

\subsection{Discrete Optimization}
After localizing a transformed trajectory $Q'$ based on time-unknown and time-known anchors, we project $Q'$ onto the valid space $\Omega\setminus \Oobs$ while avoiding the collision with the obstacle regions in $\Oobs$ by a discrete optimization based on a Viterbi algorithm. As illustrated in Fig.~\ref{fig:optim}, we first construct a 2D undirected graph where vertices $\mathcal{V} \subset \Omega$ are the concatenation of trajectory $Q'$ and random positions sampled uniformly from the valid space $\Omega \setminus \Oobs$. For each pair of vertices $(v, v') \in \mathcal{V}^2$, we validate if the straight line segment between them does not collide with any obstacle region in $\Oobs$, and span edges $\mathcal{E}\subset \mathcal{V}^2$ between valid vertex pairs. 

On the constructed graph, a trajectory can be represented by a sequence of connected vertices $\zeta=(v_1,\dots,v_T)\in \mathcal{V}^T,\; (v_{t-1}, v_t)\in\mathcal{E}$. We define a cost $c(\zeta, Q')$ for each trajectory as follows:
\begin{equation}
    c(\zeta, Q')=\sum_{t=1}^{T}\|v_t - q'_t\|_2 + \beta\sum_{t=2}^{T}\|v_{t} - v_{t-1}\|_2,
    \label{eq:viterbi}
\end{equation}
where $\|\cdot\|_2$ is the L2 norm and $\beta$ is a hyperparameter. The first unary term ensures trajectories stay as close to the original one $Q'$ as possible within the valid space, while the second pairwise term encourages them to be shorter. Viterbi algorithm can efficiently find the lowest-cost trajectory on the given graph, which we refer to as the final output.

\begin{figure}[t]
    \centering
\includegraphics[width=\linewidth]{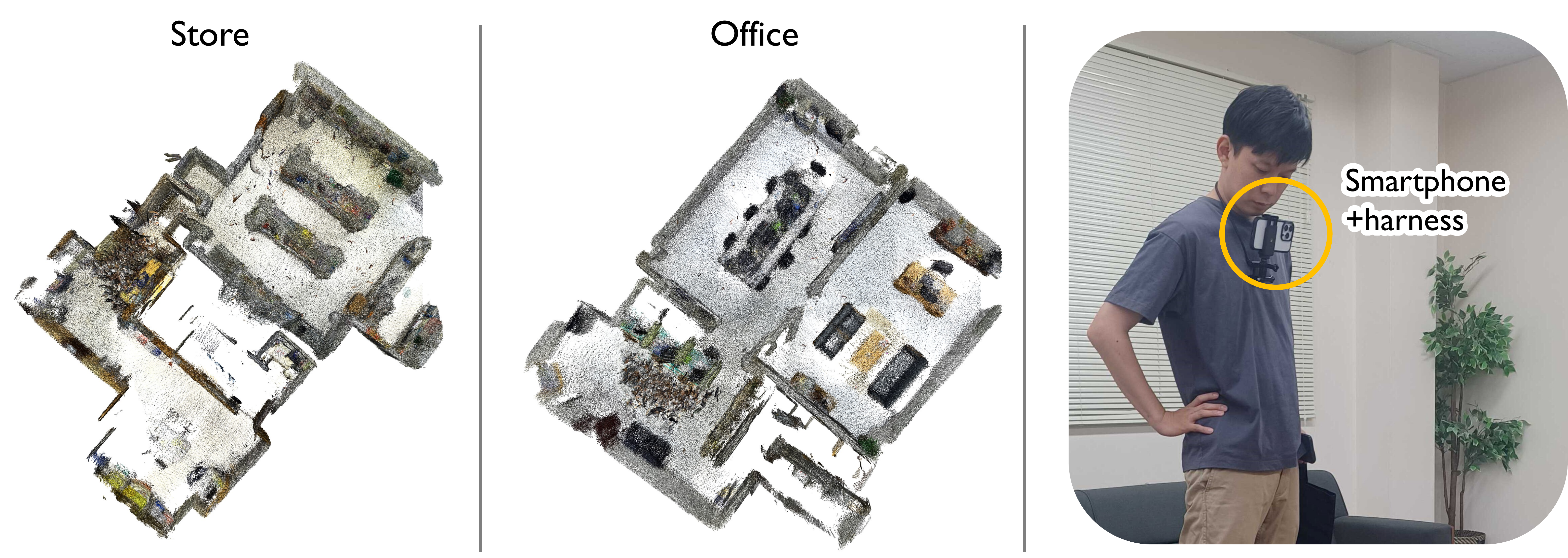}
    \caption{Data collection environments.}
    \label{fig:env}
\end{figure}

\section{Experiment}
\label{sec:user_study}
We systematically evaluate the performance of the \method system on five diverse environments: two from our new dataset and three from a public dataset~\cite{herath2022niloc}.

\subsection{Data Collection}
As shown in Fig.~\ref{fig:env}, we collected new data of people navigating two indoor environments: (a) \emph{store}, a retail store consisting of a selling area with multiple shelves and a backyard, and (b) \emph{office}, a travel agency office consisting of a hallway and two rooms with multiple desks. Both environments were approximately 100 m$^2$. 

\paragraph{Systems} We used two smartphones (iPhone 14 Pro) as wearable sensors to track participant's motion in two distinct ways. One smartphone (smartphone A) was hung from the neck using a harness and loosely fixed to ensure the camera was always horizontal and facing forward (see Fig.~\ref{fig:env} right). We performed visual-LiDAR SLAM~\cite{labbe2019rtab} to obtain a ground-truth trajectory, as done in prior work~\cite{yan2017ridi,yan2020ronin,herath2022niloc}, at a frequency of 5 Hz. The other smartphone (smartphone B) was either held in hand, placed in a pocket, or put in a bag, as per the experimenter's instructions. We recorded 3-degree-of-freedom (DoF) accelerations, 3-DoF angular rates, and 3D orientations of both devices at 50 Hz using a smartphone app~\cite{sensorlogger}. The timestamps between SLAM trajectories and IMU data of the two smartphones were synchronized based on the NTP server.

\paragraph{Procedure}
We recruited 15 participants (10 males and 5 females) who were all familiar with operating a smartphone and had no difficulty walking at a normal speed. We asked the participants to mimic shopping activities; they were asked to visit four specified shelves in the environment. Upon arriving at each shelf, they paused for a few seconds to view a poster displayed there. Participants were permitted to walk at any pace and pause whenever they wished, as long as their behavior mimicked typical indoor walking activity. After visiting all the shelves, they returned to the starting point. This process was repeated three times while varying the way smartphone B was carried (\ie, held in hand, placed in a pocket, or put in a bag), resulting in a single session of approximately three minutes. After a short break and device re-calibration, participants conducted the next session with a different set of four locations. Each participant completed three sessions for both store and office environments. We also gathered additional data by asking participants to freely walk around the environments for training neural inertial navigation models. In total, we collected about 22.8 hours of recording from the 15 participants.

\paragraph{Data Preprocessing}
To generate relative motion trajectories $Q$, we adopted RoNIN~\cite{yan2020ronin}, a popular neural inertial navigation method that estimates velocity vectors from IMU data. We split 15 participants into 10 training, 2 validation, and 3 test subsets, and trained a RoNIN-Resnet model for each environment independently. The best checkpoints were swapped between environments to ensure no overlaps between participants and environments in the training and test subsets. The four pre-determined shelves were treated as time-unknown anchors, $\Xtu$, while in practical retail setups they can easily be acquired from the store map and purchase records. Furthermore, we split each session data into three sub-sequences based on how smartphone B was held, and treated the first and last ground-truth locations with timestamps for each sub-sequence as time-known anchors, $\Xtk$.  The obstacle layout $\Oobs$ was estimated from the point cloud obtained via SLAM, although it can be identified from a store map in practice.

\subsection{Evaluation Procedure}
\paragraph{Implementation Details for \method}
In the transformation network, $\mathrm{MLP}$ consists of two layers with the channel dimensions $64$ and $128$. For the attention layer ($\mathrm{attn}$), feature dimension $d$ of $W_\mathrm{q}, W_\mathrm{k}, W_\mathrm{v}$ were all set to $d=32$. $\lambda, \beta$ in Eq.~(\ref{eq:objective}) and Eq.~(\ref{eq:viterbi}) were set to $\lambda=\beta=0.01$. The temperature parameter, $\tau$ in Eq.~(\ref{eq:weight}), was initially set to $\tau=100$ and then optimized through gradient descent. The optimization of the transformation network was done for $1000$ iterations using the Adam optimizer~\cite{kingma2014adam} with a learning rate of $0.001$. For the discrete optimization procedure, $1000$ vertices were sampled from the uniform distribution to construct a graph.

\paragraph{Baseline Methods}
Our focus in this experiment is on the fusion of relative motion trajectories from IMU data, time-unknown/known anchors, and the obstacle layout for indoor trajectory estimation. This is a novel problem setup, where it is not feasible to adopt existing fusion approaches for IMU and wireless anchors~\cite{kim2021deep,li2015bluetooth,venkatnarayan2019enhancing,zhou2023gps,herath2021fusion} due to the lack of time information for most of our anchors. Instead, we compare our approach with the following relevant methods, each utilizing different sets of input data modalities.
\begin{itemize}
    \item \textbf{TSP Solver}: This baseline solves a traveling salesman problem (TSP) to determine the shortest route that connects the two time-known anchors while visiting all the time-unknown anchors. Subsequently, we linearly interpolate the anchors to estimate a dense trajectory of length $T$ with a constant velocity. Finally, we apply the same Viterbi-based discrete optimization procedure as \method to project the trajectory onto the valid space.
    \item \textbf{RoNIN}~\cite{yan2020ronin}: This is a popular machine-learning-based inertial navigation method that estimates relative motion trajectories from IMU data, which is also employed in our system to provide initial relative trajectories. We manually geo-localize the initial locations of the trajectories to the start locations provided as time-known anchors. Note that we refrained from applying the discrete optimization procedure used in the proposed method as well as the other baselines as doing so did not improve performances.
    \item \textbf{CPD Solver}: This baseline estimates the alignment between relative trajectories from IMU and time-unknown anchors through coherent point drift (CPD) point-set registration. We evaluated three variants: nonrigid CPD~\cite{myronenko2010point}, constrained CPD~\cite{golyanik2016extended}, and Bayesian CPD~\cite{hirose2020bayesian}, which are all implemented in~\cite{probreg}. The constrained CPD uses time-known anchors as the constraints from known point-to-point correspondences. All these methods are followed by the same discrete optimization procedure using the obstacle layout as \method, thus using exactly the same set of input data.
\end{itemize}

\begin{table}[t]
    \caption{\textbf{Comparisons of average positional error (m) on our new dataset}. The \textbf{best} and \underline{second-best} methods are highlighted in bold and underlined, respectively.}
    \label{tab:result_patrol}
    \centering
    \scalebox{0.7}{
    \begin{tabular}{lccccccccc}
    \toprule 
    & \multicolumn{4}{c}{Store} & \multicolumn{4}{c}{Office} \\
    \cmidrule(lr){2-5}  \cmidrule(lr){6-9} 
     & neck & pocket & hand & bag & neck & pocket & hand & bag \\
    \midrule
     TSP & 3.08 &  2.60 &  3.16 & 3.26 & 3.76 & 3.99 & 3.63 & 3.65 \\
     RoNIN~\cite{yan2020ronin}  & 3.09 & 3.27 & 4.23 & 4.74 & 3.50 & 4.48 & 5.09 & 6.41 \\
     Nonrigid CPD~\cite{myronenko2010point} &  \underline{2.11} &  \underline{1.51} &  \underline{2.90} & \underline{3.29} & \underline{1.47} & \underline{2.26} & \underline{2.46} & \underline{3.36} \\
     Constrained CPD~\cite{golyanik2016extended} &  4.11 &  3.74 & 4.58 & 5.42 & 3.83 &  3.55 & 3.38 & 5.12 \\
 Bayesian CPD~\cite{hirose2020bayesian} &   3.17 &  2.26 &  3.48 & 3.38 & 2.18 &  2.71 &  3.09 & 3.58 \\
 \midrule
 \textbf{\method (Ours)} &         \textbf{1.25} & \textbf{0.82} & \textbf{2.43} &  \textbf{2.45} & \textbf{0.70} &            \textbf{1.17} & \textbf{1.29} &          \textbf{2.73} \\
    \bottomrule 
    \end{tabular}
    }
\end{table}

\begin{figure}[t]
    \centering
\includegraphics[width=\linewidth]{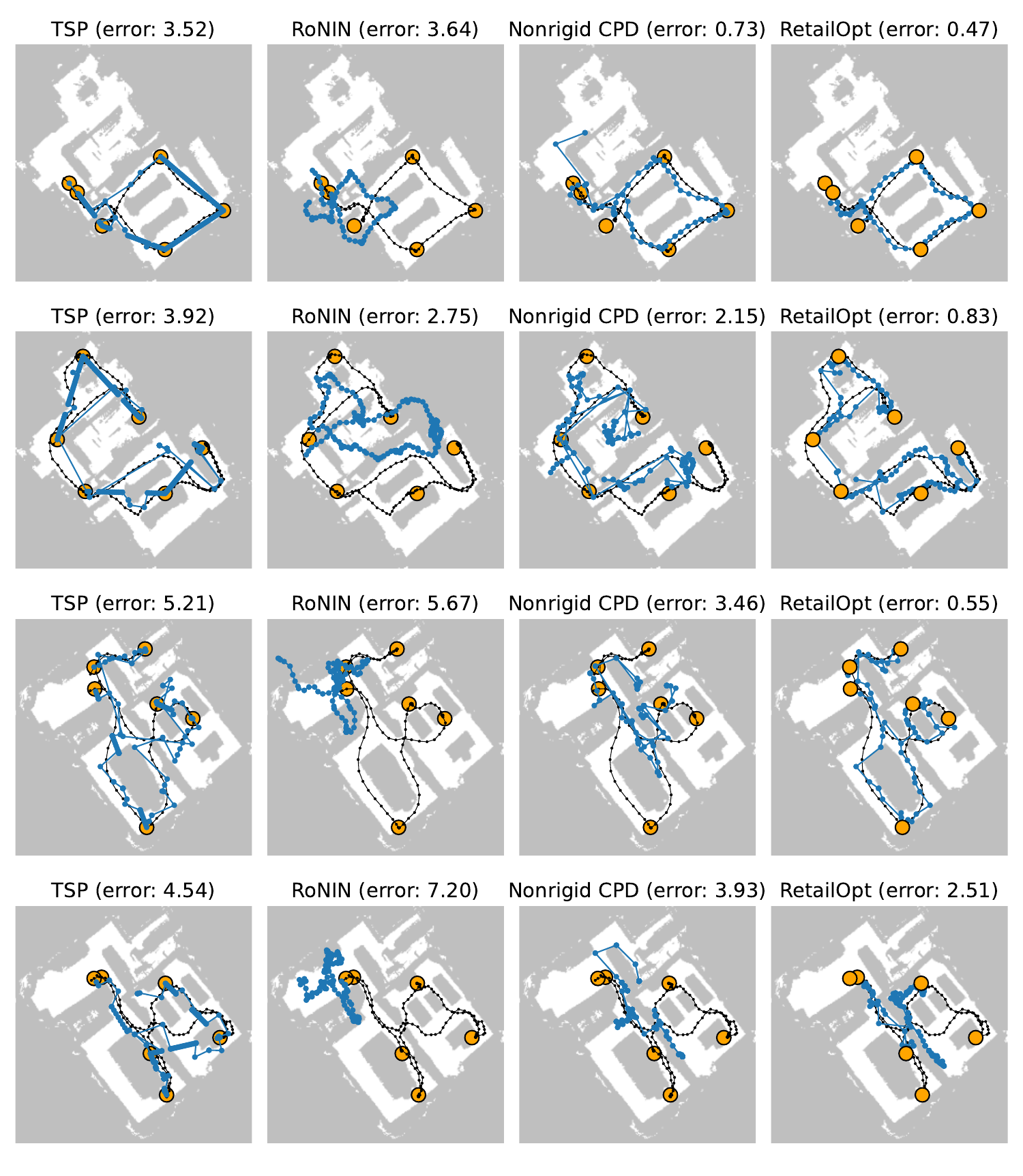}
\vspace{-2em}
    \caption{\textbf{Results on our new dataset.} The estimated and ground-truth trajectories are visualized with blue and black lines, respectively. Obstacle regions are colored in gray, and anchors are annotated with orange circles.}
\vspace{-1em}
    \label{fig:result_patrol}
\end{figure}

\subsection{Results on Our New Dataset}
\paragraph{Quantitative Results}
We evaluated the positional error between the estimated and ground-truth trajectories, which is reported in meters and averaged over time steps at 2 Hz. We hypothesize that the performance of \method would depend on the quality of the initial relative trajectories provided by RoNIN. As one of the sources for erroneous outputs from RoNIN is the variability of smartphone poses, we systematically calculated the positional errors separately for each phone's handling style: neck (from smartphone A's IMU), pocket, bag, and hand (from smartphone B). As shown in Tab.~\ref{tab:result_patrol}, \method estimates participant's trajectories more accurately than baseline methods. The performances of TSP and RoNIN were limited due to their access being restricted to only anchors and obstacle layouts (in the case of TSP) or dense motion trajectories (for RoNIN). Among the CPD family, nonrigid CPD proved to be the most robust, suggesting that point-to-point constraints and Bayesian inferences for the other CPD methods were not particularly effective in further enhancing the estimation accuracy. For the \method, the average positional error is less than 1.3 m particularly when the smartphone was hung from the neck or in a pocket. It can process trajectories of approximately 600
seconds in less than 2 seconds on a single GPU (NVIDIA Tesla T4). Note that it was challenging for all the methods and environments to accurately estimate trajectories when the smartphone was put freely in a bag, due to the degraded accuracy of RoNIN that provided initial trajectories. Using IMUs in other devices tightly attached to people, such as watches and earphones, may provide more stable performance.

\paragraph{Qualitative Results}
Fig.~\ref{fig:result_patrol} visualizes some results from TSP, RoNIN, nonrigid CPD, and \method. \method estimates trajectories accurately even for challenging cases where the other baselines failed to work (the third example), but could provide inaccurate results if the RONIN's prediction is completely corrupted (the bottom example). Nonrigid CPD and TSP baselines could work on simple trajectories, but their performances significantly degraded for complex problems. We found that there were often significant spatial overlaps between \method's trajectories and ground-truth ones. This indicates that the main source of errors in \method comes from temporal discrepancy rather than spatial one.

\begin{table}[t]
    \caption{\textbf{Comparisons of average positional error (m) on Inertial Localization Dataset~\cite{herath2022niloc}}. The \textbf{best} and \underline{second-best} methods are highlighted in bold and underlined, respectively.}
    \label{tab:result_niloc}
    \centering
    \scalebox{0.65}{
    \begin{tabular}{lccccccccc}
    \toprule 
    & \multicolumn{3}{c}{University A} & \multicolumn{3}{c}{University B} & \multicolumn{3}{c}{Office} \\
    \cmidrule(lr){2-4}  \cmidrule(lr){5-7}  \cmidrule(lr){8-10} 
    Distance traveled (m) & 100 & 200 & 400 & 100 & 200 & 400 & 100 & 200 & 400 \\
    \midrule
    TSP & 12.95 & 17.46 & 22.55 & 11.88 & 20.86 & 20.54 & 5.65 & 6.69 & 10.53 \\ 
    RoNIN~\cite{yan2020ronin} &  7.18 & 7.42 & \underline{8.19} & 5.90 & 8.35 & 20.91 & 2.70 & 3.74 & 6.51 \\
    Nonrigid CPD~\cite{myronenko2010point} & \underline{5.94} & 7.37 & 8.39 & 2.79 & 12.81 & 14.61 & 4.45 & 4.74 & 5.06\\
    Constrained CPD~\cite{golyanik2016extended} & 4.35 & \underline{5.45} & 8.85 & \underline{1.94} & 11.94 & 11.91 & 3.01 & 3.98 & 5.78 \\
    Bayesian CPD~\cite{hirose2020bayesian} &  7.59 & 10.95 & 12.07 & 4.67 & 16.30 & 17.92 & 5.60 & 5.88 & 5.74 \\
    NILoc~\cite{herath2022niloc} (within-envs) & 21.04 & 20.67 & 20.07 & 5.61 & \underline{4.65} & \textbf{6.19} & \underline{2.62} & \textbf{1.94} & \textbf{2.06} \\
    NILoc~\cite{herath2022niloc} (cross-envs) & 45.25 & 41.82 & 42.16 & 53.60 & 55.77 & 52.76 & 16.56 & 14.71 & 13.56 \\
    \midrule
    \textbf{\method (Ours)} & \textbf{1.35} & \textbf{1.92} & \textbf{2.62} & \textbf{1.65} & \textbf{4.33} & \underline{8.93} & \textbf{1.45} & \underline{2.16} & \underline{3.38} \\
    \bottomrule
    \end{tabular}
    }
\end{table}

\begin{figure}[t]
    \centering
\includegraphics[width=\linewidth]{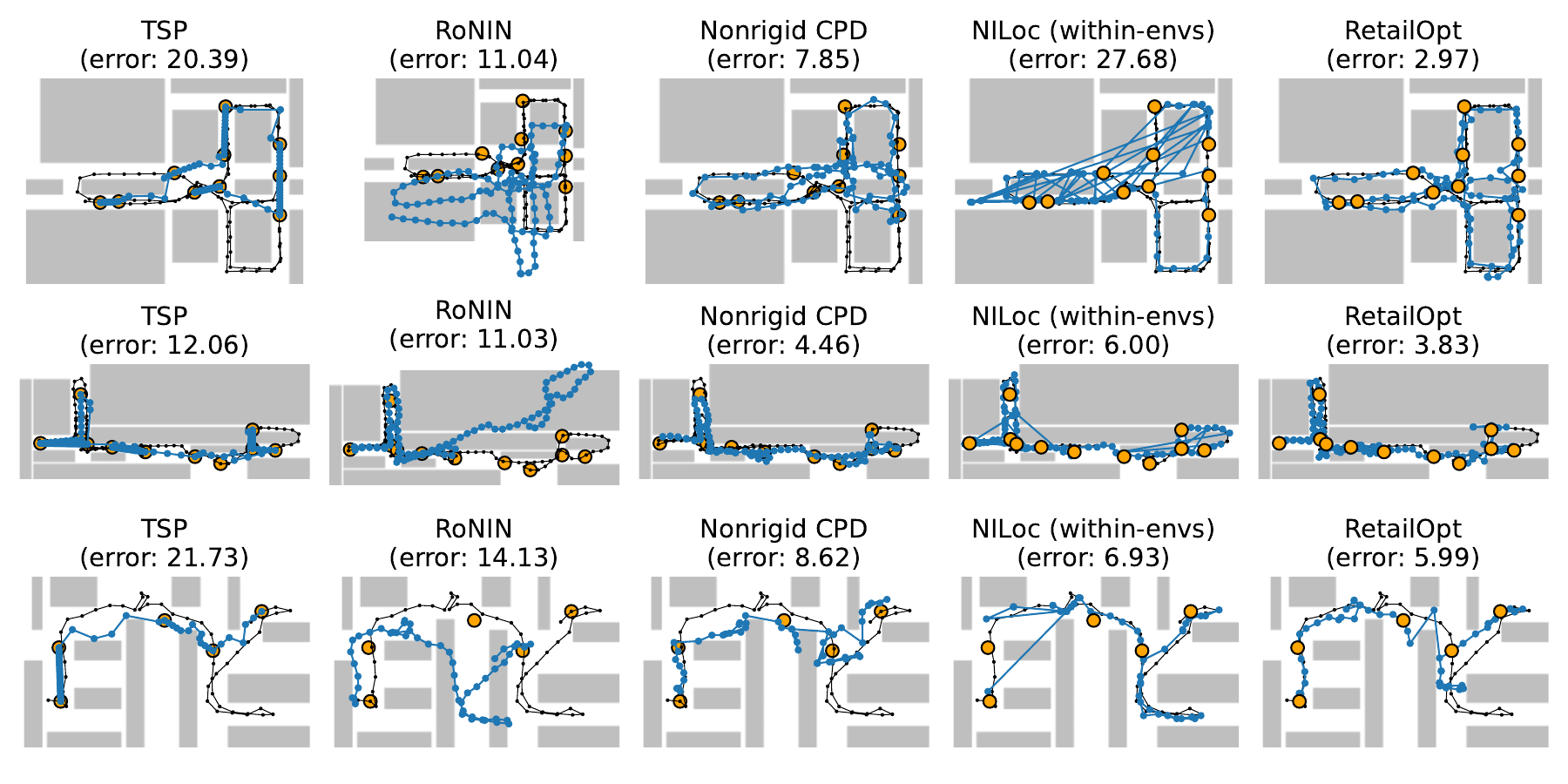}
    \caption{Results on Inertial Localization Dataset. The estimated and ground-truth trajectories are visualized with blue and black lines, respectively. Obstacle regions are colored in gray, and anchors are annotated with orange circles.}
    \label{fig:result_niloc}
\end{figure}

\balance
\subsection{Results on Inertial Localization Dataset}
\label{sec:experiment}
\paragraph{Data Preprocessing}
In addition to our new dataset, we utilized the Inertial Localization Dataset~\cite{herath2022niloc}, which includes over 50 hours of IMU data recorded in two much larger outdoor environments (University A: 62.8$\times$84.4 m$^2$; University B: 57.6$\times$147.2 m$^2$) and one indoor environment (Office: 38.4 $\times$11.2 m$^2$). Participants were instructed to navigate the environment while handling a pair of smartphones. One smartphone was held straight in one hand to record ground-truth trajectories via SLAM. The other smartphone was used freely, such as being held in the hand, used for a phone call, or stored in a pocket, although specific ways to hold it were not annotated. To make the dataset compatible with our problem setup, we performed the following pre-processing:
\begin{itemize}
\item \textbf{Obstacle layout}: For each environment, we aggregated all the ground-truth trajectories to identify the valid space (\ie, walkable area) following the instructions of \cite{herath2022niloc}. We then manually annotated rectangular obstacle regions to serve the obstacle layout $\Oobs$.
\item \textbf{Anchors}: Additionally, we created some candidates for anchors by thinning the valid space and performing corner detection. As shown in Fig.~\ref{fig:result_niloc}, detected corners are located around the intersections or middle of the corridors. For each data sample, we selected a part of the anchors that were within approximately 1 m around the ground-truth trajectory, and used them as variable numbers of time-unknown anchors, $\Xtu$. We also used the starting and ending points of each trajectory as time-known anchors, $\Xtk$.
\end{itemize}

\paragraph{Results}
Tab.~\ref{tab:result_niloc} presents the quantitative comparisons between \method and the baselines. Given the longer ground-truth trajectories in larger environments and due to the lack of annotations for how smartphones were hold, we instead reported average positional errors for several thresholds of distances being traveled. \method consistently outperformed the other baselines. Furthermore, we evaluated NILoc~\cite{herath2022niloc}, a state-of-the-art inertial localization method released with the dataset. \textbf{NILoc (within-envs)} model was trained on the same environment, while \textbf{NILoc (cross-envs)} model was trained on another environment. Comparing these two variants, we found that NILoc performed effectively only for the within-envs condition, demonstrating the necessity of a site survey for a large-scale data collection at each environment. In contrast, our proposed system does not require additional training data to be deployed in new environments. Fig.~\ref{fig:result_niloc} shows some qualitative examples.

\subsection{Discussions}
\label{sec:discussions}

\paragraph{Limitations}
Our system is currently built upon several assumptions, which could limit its potential use-cases. First, we presume a one-to-one match between items and their actual locations in environments. If identical items are stored on multiple shelves, this could obscure the locational cues from time-unknown anchors. Second, our approach assumes the store map and purchase records are provided by a single store. Estimating trajectories across multiple stores next to each other requires a unified data analysis platform that can access information from every store, certainly with different owners. While this is technically feasible, it implies the use of one retailer's store information to improve the quality of trajectory data provided to another retailer, necessitating an appropriate incentive design and data protection framework.

\paragraph{Applications to Other Domains} 
Besides the retail applications, the proposed system is also applicable to other scenarios such as office maintenance and museum visits~\cite{kontarinis2017towards}. Many facilities have and can provide detailed map data, and it is now becoming common for people to install smartphone apps to record what items they have interacted with or which events they have experienced (\eg, a list of locations to be inspected for office maintenance tasks, or that of artworks experienced in the museum). These data are analogous to the store map and purchase records for the retail scenario.
\section{Conclusion}
\label{sec:conclusion}
We have introduced \method, an opt-in and easy-to-deploy indoor trajectory estimation system designed for retail environments. The system exploits built-in retail facilities, namely the store map and purchase records, to extract location anchors where the target customer has or has not visited. Smartphones are utilized as a wearable sensor to estimate relative motion trajectories of the customer via neural inertial navigation. Continuous and discrete optimization uses the anchors to localize the trajectories in-store. Systematic experiments have demonstrated the effectiveness of our system based on our new data collection from retail environments and a public inertial localization dataset.
\paragraph{Acknowledgements.}
This work was supported by JSPS KAKENHI Grant Number 23K24831.

\clearpage

\bibliographystyle{ACM-Reference-Format}
\bibliography{ref}

\end{document}